\pdfoutput=1

\documentclass[11pt]{article}

\usepackage[preprint]{acl}

\usepackage{times}
\usepackage{latexsym}
\usepackage{algorithm}
\usepackage{algpseudocode}
\usepackage{amsmath}
\usepackage{amsfonts}  
\usepackage{amssymb}
\usepackage{multirow}
\usepackage{graphicx}
\usepackage{subcaption}
\usepackage{makecell}
\usepackage{booktabs}
\usepackage{array}
\usepackage{url}
\usepackage{dsfont}
\usepackage{xspace}
\usepackage{tcolorbox}
\usepackage{pifont}
\usepackage{xcolor}
\usepackage[T1]{fontenc}
\usepackage{booktabs} 
\usepackage{graphicx} 
\usepackage{amssymb}  
\usepackage{siunitx}  
\usepackage{array}    
\usepackage{soul}     

\usepackage[utf8]{inputenc}

\usepackage{microtype}

\usepackage{inconsolata}

\usepackage{graphicx}

\tcbuselibrary{listings, breakable, skins, theorems}
\newtcolorbox{prompt}[1]{
    enhanced,
    drop shadow=black!5!white,
    left=4mm,
    right=4mm,
    top=1mm,
    bottom=1mm,
    boxsep=0mm,
    rounded corners,
    title=#1,
    fontupper=\normalsize\linespread{1}\fontfamily{lmr}\selectfont,
}

\title{CC-GSEO-Bench: A Content-Centric Benchmark for   \\Measuring Source Influence in Generative Search Engines}

\author{
\textbf{Qiyuan Chen}$^{1,2}$\thanks{\, Equal Contribution.},
\textbf{Jiahe Chen}$^{1}$\footnotemark[1],
\textbf{Hongsen Huang}$^{2}$\footnotemark[1],
\textbf{Qian Shao}$^{1}$,
\textbf{Jintai Chen}$^{3}$,\\
\textbf{Renjie Hua}$^{2}$,
\textbf{Hongxia Xu}$^{1}$,
\textbf{Ruijia Wu}$^{4}$,
\textbf{Chuan Ren}$^{2}$\footnotemark[2],
\textbf{Jian Wu}$^{1}$\thanks{\, Corresponding Author.}\\
$^{1}$Zhejiang University, Zhejiang, China
$^{2}$Soochow Securities, Jiangsu, China\\
$^{3}$HKUST(GZ), Guangdong, China
$^{4}$Shanghai Jiaotong University, Shanghai, China\\
\texttt{qiyuanchen@zju.edu.cn}
}

\begin{document}
\maketitle

\begin{abstract}
Generative Search Engines (GSEs) synthesize conversational answers from multiple sources, weakening the long-standing link between search ranking and digital visibility.
This shift raises a central question for content creators: How can we reliably quantify a source article's influence on a GSE's synthesized answer across diverse intents and follow-up questions?
We introduce \textbf{CC-GSEO-Bench}, a content-centric benchmark that couples a large-scale dataset with a creator-centered evaluation framework.
The dataset contains over 1{,}000 source articles and over 5{,}000 query--article pairs, organized in a one-to-many structure for article-level evaluation.
We ground construction in realistic retrieval by combining seed queries from public QA datasets with limited synthesized augmentation and retaining only queries whose paired source reappears in a follow-up retrieval step.
On top of this dataset, we operationalize influence along three core dimensions, \textbf{Exposure}, \textbf{Faithful Credit}, and \textbf{Causal Impact}, and two content-quality dimensions, \textbf{Readability and Structure} and \textbf{Trustworthiness and Safety}.
We aggregate query-level signals over each article's query cluster to summarize influence strength, coverage, and stability, and empirically characterize influence dynamics across representative content patterns.
\end{abstract}

\section{Introduction}

For decades, the dominant model for information access has been that of \emph{Traditional Search Engines} (TSEs), which return a ranked list of hyperlinks for users to navigate~\citep{schwartz1998web,arasu2001searching,schutze2008introduction}.
The recent advent of powerful Large Language Models (LLMs) has catalyzed a paradigm shift~\citep{ouyang2022training,achiam2023gpt}, giving rise to \emph{Generative Search Engines} (GSEs), such as Perplexity and Copilot.
Rather than merely retrieving documents, GSEs synthesize information from multiple sources into a single conversational response, transforming the information-seeking process from navigation to inquiry~\citep{aggarwal2024geo,allan2024future}.

\begin{figure}
    \centering
    \includegraphics[width=\linewidth]{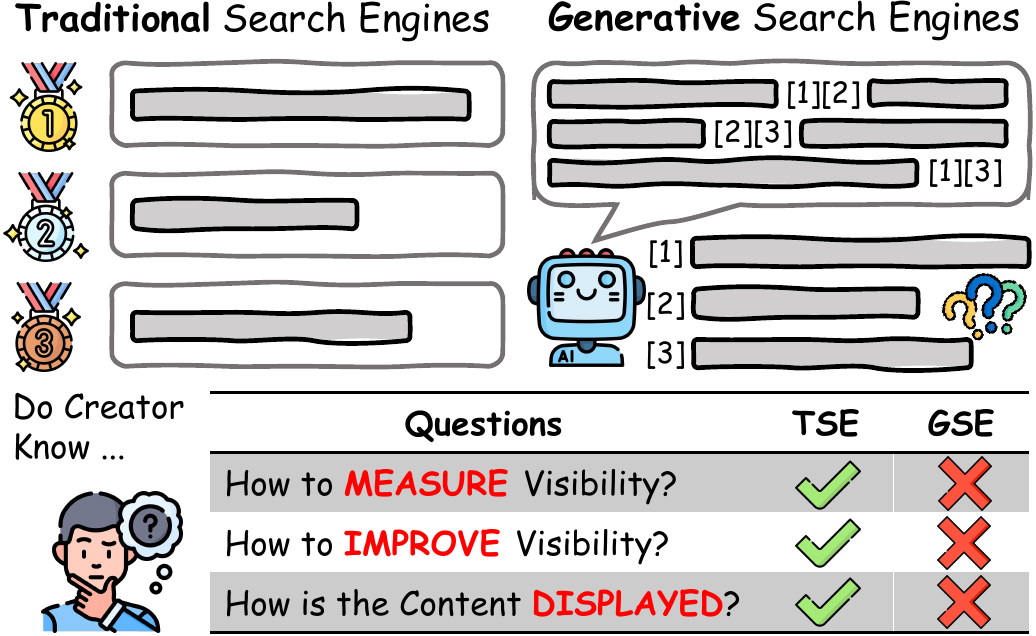}
    \caption{Shift from TSEs with ranked visibility to GSEs with synthesized answers, illustrating the new uncertainties for creators in measuring and influencing content display.}
    \label{fig:compare_search}
    \vspace{-6mm}
\end{figure}

This shift from ranked retrieval to generative synthesis is fundamentally redefining digital visibility.
In the TSE era, visibility is largely governed by a source's position on the Search Engine Results Page (SERP), making rank-oriented Search Engine Optimization (SEO) the dominant optimization target~\citep{davis2006search}.
In contrast, GSEs supplant the SERP with a synthesized answer that may include only a subset of sources, thereby weakening the connection between ranking and visibility~\citep{aggarwal2024geo,nestaas2024adversarial}.
As shown in Figure~\ref{fig:compare_search}, the central challenge for content creators shifts from ``ranking higher'' to achieving measurable influence on the synthesized answer.
This raises a critical research question: \emph{How can we reliably quantify a source article's influence on a GSE's synthesized answer across the diverse ways users ask about the same topic?}

To answer this question, we first ask a more fundamental one: What does influence mean from a content creator's perspective?
We argue that meaningful influence goes beyond performance on any single query.
Content creators rarely care about optimizing for a single query; instead, they want their article's core facts and insights to consistently shape users' understanding across a cluster of related intents, paraphrases, and follow-up questions.
Accordingly, any meaningful measurement must go beyond isolated query-level snapshots and aggregate influence at the article level over many potential user interactions~\citep{liu2023evaluating}.

Guided by this creator-centered notion of influence, we introduce \textbf{CC-GSEO-Bench}, a unified content-centric benchmark that combines a large-scale dataset with a creator-centered evaluation framework.
The dataset organizes each instance around a source article and its associated query cluster in a one-to-many structure, enabling article-level evaluation across diverse user intents.
In total, CC-GSEO-Bench contains over 1{,}000 source articles and over 5{,}000 query--article pairs.
To improve ecological validity, we ground data construction in realistic retrieval.
We begin with seed queries drawn primarily from public QA datasets and additionally synthesize a small set of queries to broaden intent coverage.
Candidate sources are obtained via web search, and we retain a query only if the paired source reappears in a follow-up retrieval step.

Built on top of this dataset, we operationalize creator-centered influence along three core dimensions and two content-quality dimensions.
The core influence dimensions are \textbf{Exposure}, \textbf{Faithful Credit}, and \textbf{Causal Impact}.
Exposure captures whether a source is included in the synthesized answer and whether its presence is visible and salient to users.
Faithful Credit evaluates whether claims attributed to the source are actually supported by the source and whether the source's meaning is preserved without distortion.
Causal Impact measures the degree to which the source changes answer quality by comparing answers generated with the source present versus with the source removed from the retrieval context.
To support article-level analysis, we aggregate these query-level signals over the query cluster associated with the same article, summarizing influence in terms of strength, coverage, and stability across varied intents and paraphrases.
Finally, we include two content-quality dimensions that help interpret and diagnose influence outcomes.
\textbf{Readability and Structure} assesses the clarity and organization of the source itself, while \textbf{Trustworthiness and Safety} examines whether the source meets basic standards of reliability and avoids deceptive or harmful content.
Together, the dataset and these dimensions constitute CC-GSEO-Bench as a unified benchmark for measuring and analyzing source influence in generative search.

Our contributions are threefold.
First, we introduce CC-GSEO-Bench, a large-scale content-centric benchmark that pairs each source article with a query cluster, enabling systematic article-level evaluation of influence across diverse user intents.
Second, we propose a creator-centered evaluation framework that operationalizes influence with three core dimensions (Exposure, Faithful Credit, Causal Impact) and two content-quality dimensions (Readability and Structure, Trustworthiness and Safety), together with aggregation metrics that summarize influence strength, coverage, and stability over a query cluster.
Third, we conduct an empirical study on CC-GSEO-Bench to characterize influence dynamics and trade-offs across representative content patterns, providing actionable insights for creators and a shared measurement foundation for future research in generative search engine optimization.

\begin{figure*}[ht]
    \centering
    \includegraphics[width=0.98\linewidth]{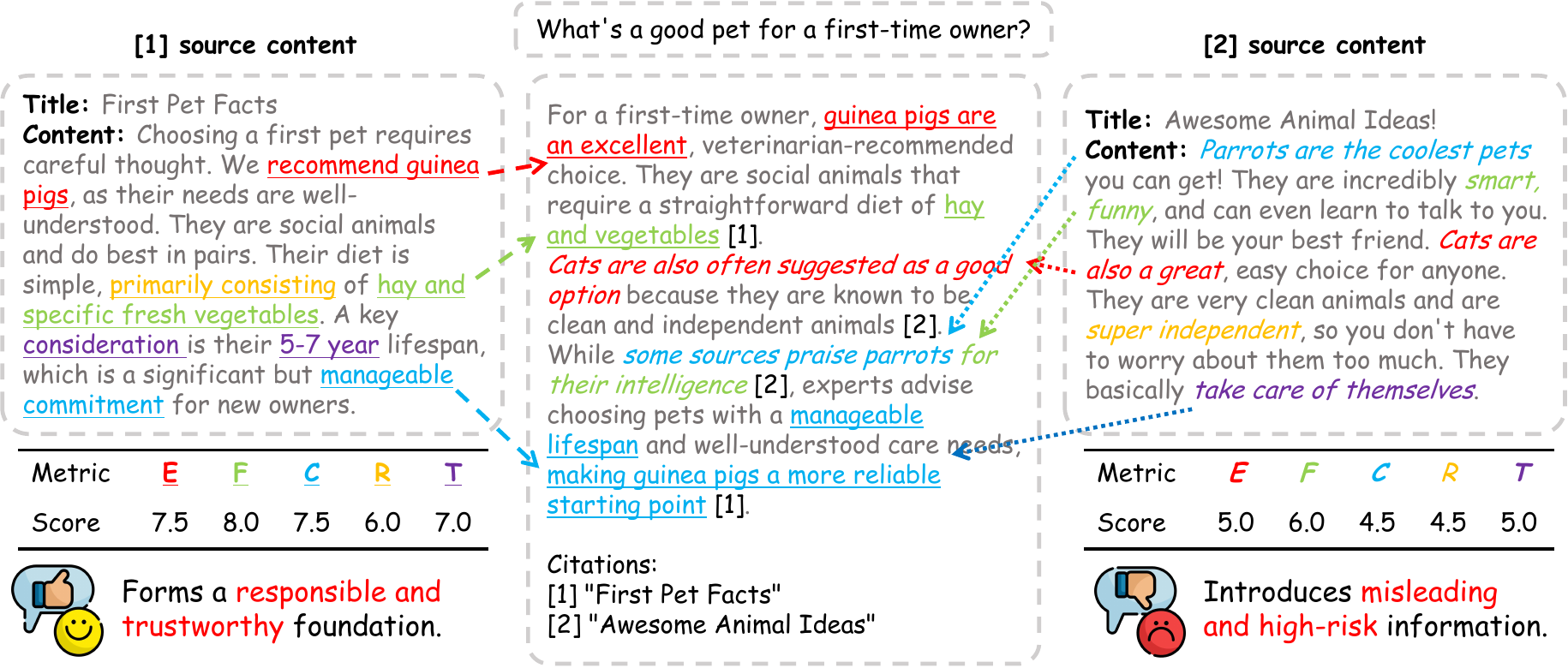}
    \caption{Illustration of the CC-GSEO-Bench framework. We quantify influence through five dimensions: Exposure (E), Faithful Credit (F), Causal Impact (C), Readability (R), and Trustworthiness (T). In this example, Source 1 demonstrates high Readability and Trustworthiness which translates into dominant Exposure and Causal Impact as its logical framework shapes the final recommendation. Conversely, Source 2 exhibits low Trustworthiness and fails to positively influence the synthesized answer despite being retrieved.}
    \label{fig:gseo-framework}
    \vspace{-4mm}
\end{figure*}

\section{CC-GSEO-Bench: A Content-Centric Benchmark for Generative Search Influence}
\label{sec:benchmark}

\subsection{Task Formulation: Measuring Influence in Generative Search}
\label{sec:task}

Generative Search Engines (GSEs) answer a user query by synthesizing information from multiple retrieved sources. In this setting, a document being retrieved does not imply that it will be reflected in the final response. CC-GSEO-Bench therefore centers on \emph{source influence}, namely the extent to which a particular source document shapes the synthesized answer under realistic retrieval contexts and diverse query formulations.

\paragraph{Content-centric evaluation unit.}
Let $\mathcal{S}=\{S_1,\dots,S_M\}$ denote a set of $M$ source articles. CC-GSEO-Bench adopts a content-centric structure in which each article $S_a$ is paired with an article-conditioned query set $\mathcal{Q}_a=\{q_{a,1},\dots,q_{a,N_a}\}$. This design matches a realistic creator-side objective: a single piece of content should remain influential across a family of plausible user intents and paraphrases related to the same topic, rather than being tuned to a single canonical query. It also enables article-level aggregation and robustness analysis across query variants.

\paragraph{Offline generative search simulation.}
To ensure reproducibility without depending on proprietary GSE APIs, we define an offline simulator consisting of a retriever $\mathsf{R}$ and a generator $\mathsf{G}$. For each pair $(S_a,q_{a,j})$, we form a retrieval context $\mathcal{C}^{+}_{a,j}$ that contains the target article $S_a$ and additional retrieved documents:
\begin{equation}
\mathcal{C}^{+}_{a,j} = \big[d^{(1)}_{a,j},\dots,d^{(K)}_{a,j}\big],\quad S_a \in \mathcal{C}^{+}_{a,j}.
\end{equation}
The generator $\mathsf{G}$ is prompted to answer the query using the provided context, where the context documents are presented as a numbered reference list. The generated answer is required to use explicit source markers that refer to items in $\mathcal{C}^{+}_{a,j}$ (e.g., \texttt{[1]}, \texttt{[2]}), so that downstream evaluation can attribute visibility and credit to individual sources:
\begin{equation}
A^{+}_{a,j} = \mathsf{G}\!\left(q_{a,j},\mathcal{C}^{+}_{a,j}\right).
\end{equation}

\paragraph{Counterfactual answers for incremental contribution.}
A central aspect of influence is what changes when a source is present versus absent under the same background evidence. We therefore construct a counterfactual context by removing the target source while keeping all other retrieved documents unchanged:
\begin{equation}
\mathcal{C}^{-}_{a,j} = \mathcal{C}^{+}_{a,j} \setminus \{S_a\}, \qquad
A^{-}_{a,j} = \mathsf{G}\!\left(q_{a,j},\mathcal{C}^{-}_{a,j}\right).
\end{equation}
Intuitively, $A^{+}_{a,j}$ and $A^{-}_{a,j}$ answer the same query under the same retrieval environment, except that only $A^{+}_{a,j}$ can draw on $S_a$. Comparing the two answers supports a counterfactual estimate of the target source's marginal contribution beyond what other retrieved documents already provide.

\subsection{Benchmark Construction}
\label{sec:construction}

CC-GSEO-Bench is constructed through a multi-stage pipeline that collects real web articles and pairs them with validated, article-specific query sets, while preserving the retrieval contexts needed for reproducible offline evaluation. Detailed statistics on the benchmark's composition and distribution are provided in Appendix~\ref{benchmark_construction}.

We begin with a diverse pool of seed questions drawn from multiple public QA datasets, covering factual, explanatory, and multi-hop information needs. For each seed question, we apply the retriever $\mathsf{R}$ to obtain the top-$K_0$ web results and then randomly sample one result page as a candidate source article. We store the URL and raw HTML and extract the main article text by removing boilerplate such as navigation bars and repeated template blocks.

Given a cleaned article $S_a$, we use an instruction-tuned LLM to synthesize a candidate query set $\mathcal{Q}_a$ conditioned on the article content. The synthesis prompt is designed to produce questions that are answerable from the article while encouraging diversity across plausible user intents, such as definitions, comparisons, procedures, and pros and cons. We further filter near-duplicate or repetitive questions to ensure that the resulting query set reflects meaningful variations rather than paraphrases that are trivially redundant.

To ensure that query--article pairs correspond to realistic retrieval behavior, we perform a retrieval validation step. For each generated query $q_{a,j}$, we re-run the retriever $\mathsf{R}$ and keep $q_{a,j}$ only if the target article $S_a$ appears in the top-$K_0$ retrieved results. This step removes pairs that are only loosely related and unlikely to co-occur in practice, and it also strengthens the causal interpretation of counterfactual removal by ensuring that the target source is plausibly available to the generator in the factual context.

For reproducibility, we cache the retrieval lists for each $(S_a,q_{a,j})$, including URLs, titles, snippets or summaries, and the rank positions of retrieved documents. These cached contexts allow others to reconstruct identical evaluation inputs and to compare influence measurements under the same retrieval evidence, without relying on identical access to a live search API.

\section{Measuring Source Influence}
\label{sec:metrics}

\subsection{Notation}
\label{sec:notation}

For each source article $S_a$ and query $q_{a,j}$, the simulator produces an answer $A^{+}_{a,j}$ from context $\mathcal{C}^{+}_{a,j}$ and a counterfactual answer $A^{-}_{a,j}$ from $\mathcal{C}^{-}_{a,j}$. We measure influence along five dimensions, each yielding a micro-level score
\begin{equation}
d_k(S_a,q_{a,j}) \in [0,10],
\end{equation}
where larger values indicate stronger influence or better behavior under that dimension.

\subsection{Micro-Level Dimensions}
\label{sec:dimensions}

Our dimensions are designed to align with the information available in generative search outputs and with creator-facing questions about how content appears, is credited, and changes answer quality. We distinguish three answer-level dimensions, which depend on the generated responses, from two source-level dimensions, which reflect intrinsic document quality independent of query phrasing.

\subsubsection{Exposure}
\label{sec:exposure}

Exposure measures how visible and prominent the target source is in the final answer for a given query. In our implementation, Exposure is scored by a judge model that reads the user query, the generated answer $A^{+}_{a,j}$ (including citations or source markers), and a short representation of the target source consisting of its title, URL, and a bounded snippet. The snippet is taken from a provided summary when available, otherwise it is derived from the document content and truncated to a fixed maximum length. The judge assigns an integer score from 0 to 10, where low scores correspond to cases in which the source is not referenced or is only mentioned in a negligible way, and high scores correspond to cases in which the source is clearly present in salient parts of the answer and appears to be one of its main supporting sources. Exposure intentionally focuses on prominence rather than factuality or usefulness, which are captured by other dimensions.

\subsubsection{Faithful Credit}
\label{sec:faithfulcredit}

Faithful Credit measures whether the answer uses the target source accurately and without serious distortion. The judge model receives the query, the generated answer $A^{+}_{a,j}$, and the full text (or a long excerpt) of the target source. It identifies portions of the answer that appear to rely on the target source, using explicit citations or strong textual and semantic signals, and evaluates whether those statements are supported by the source and preserve its meaning. The score ranges from 0 to 10. Low scores indicate unsupported, invented, or substantially distorted attributions to the target source, while high scores indicate that attributed content is clearly grounded in the source and presented faithfully. If the answer does not meaningfully rely on the target source, the score is expected to be low.

\subsubsection{Causal Impact}
\label{sec:causalimpact}

Causal Impact measures how much the overall usefulness and correctness of the answer depends on the target source. The judge compares the paired answers $(A^{+}_{a,j},A^{-}_{a,j})$ for the same query, where $A^{+}_{a,j}$ is generated with the target source available and $A^{-}_{a,j}$ is generated after removing it. The judge assigns a 0--10 score based on how much worse the answer becomes without the target source, considering completeness, correctness, usefulness, and clarity from the user perspective. A near-zero score indicates that the two answers are effectively similar in quality, suggesting that the target source provides little marginal value under the given retrieval context. A high score indicates that removing the source causes a clear loss of key information or a noticeable degradation in answer quality.

\subsubsection{Readability \& Structure}
\label{sec:readability}

Readability \& Structure characterizes how easy the source document is to read and navigate, and how well its organization supports information extraction in generative search settings. This dimension is evaluated at the document level: the judge reads the document text and assesses clarity, logical organization, and structural cues such as headings, paragraphing, and lists. The output is a 0--10 score, where higher values indicate clearer writing and better organization. Since this property is intrinsic to $S_a$, we treat it as constant across all queries paired with the same article and reuse the computed score across the query cluster.

\subsubsection{Trustworthiness \& Safety}
\label{sec:trust}

Trustworthiness \& Safety measures whether the source document appears reliable and avoids unsafe, harmful, or misleading content based on the text itself. This dimension is also evaluated at the document level. The judge inspects the content for red flags such as fabricated-sounding claims, manipulative framing, or harmful and illegal guidance, and assigns a 0--10 score where higher values indicate more trustworthy and safer content. As with readability, this score is treated as an intrinsic property of $S_a$ and is reused across all queries associated with the same article.

\subsection{Macro-Level Aggregation across Query Variants}
\label{sec:aggregation}

Micro-level scores quantify influence for a specific query phrasing, but CC-GSEO-Bench is designed to capture behavior across query variants associated with the same source. For each dimension $k$, we aggregate query-level scores into article-level summaries and then compute benchmark-level metrics that macro-average across articles.

For article $S_a$, we compute its mean score on dimension $k$:
\begin{equation}
\mu_{a,k} = \frac{1}{N_a}\sum_{j=1}^{N_a} d_k(S_a,q_{a,j}).
\end{equation}
We then compute the benchmark-level macro-average (each article has equal weight), termed the Mean Influence Level (MIL):
\begin{equation}
\textsc{MIL}_k = \frac{1}{M}\sum_{a=1}^{M}\mu_{a,k}.
\end{equation}

To measure how broadly an article succeeds across its query set, we define a thresholded Influence Coverage metric (ICov) with threshold $\tau_k$:
\begin{equation}
\begin{aligned}
\textsc{ICov}_k(S_a) &= \frac{1}{N_a}\sum_{j=1}^{N_a} \mathbb{I}\!\left(d_k(S_a,q_{a,j}) \ge \tau_k\right), \\
\textsc{ICov}_k &= \frac{1}{M}\sum_{a=1}^{M}\textsc{ICov}_k(S_a).
\end{aligned}
\end{equation}
This metric captures the fraction of query variants for which the source achieves at least an acceptable level of influence under dimension $k$.

Finally, to quantify consistency across different phrasings and intents within the same article-conditioned query set, we compute within-article variance:
\begin{equation}
\sigma^2_{a,k} = \frac{1}{N_a}\sum_{j=1}^{N_a}\left(d_k(S_a,q_{a,j})-\mu_{a,k}\right)^2.
\end{equation}
We convert variance into a higher-is-better Influence Stability score (IStab) by normalizing with a robust upper bound $v_k$:
\begin{equation}
\begin{aligned}
\textsc{IStab}_k(S_a) &= 1 - \min\left(1,\frac{\sigma^2_{a,k}}{v_k}\right), \\
\textsc{IStab}_k &= \frac{1}{M}\sum_{a=1}^{M}\textsc{IStab}_k(S_a).
\end{aligned}
\end{equation}
Higher $\textsc{IStab}_k$ indicates that the source behaves more consistently across query variants for the same article.
\begin{table*}[t]
\centering
\caption{Performance of different GSEO strategies on \texttt{gpt-oss-120b}. We report the Mean Influence Level (MIL), Coverage (ICov), and Stability (IStab) for the primary metrics (Exposure, Faithful Credit, Causal Impact), alongside the mean scores for Readability \& Structure and Trustworthiness \& Safety. The best performance per column is highlighted in \textbf{bold}.}
\label{tab:main_results}
\resizebox{\textwidth}{!}{
\begin{tabular}{lccccccccccc}
\toprule
\multirow{2}{*}{\textbf{Method}} & \multicolumn{3}{c}{\textbf{Exposure (E)}} & \multicolumn{3}{c}{\textbf{Faithful Credit (F)}} & \multicolumn{3}{c}{\textbf{Causal Impact (C)}} & \textbf{Readability (R)} & \textbf{Trust (T)} \\
\cmidrule(lr){2-4} \cmidrule(lr){5-7} \cmidrule(lr){8-10} \cmidrule(lr){11-11} \cmidrule(lr){12-12}
 & MIL & ICov & IStab & MIL & ICov & IStab & MIL & ICov & IStab & Mean & Mean \\
\midrule
None & 5.630 & 0.333 & \textbf{0.773} & 5.747 & 0.477 & 0.691 & 5.501 & 0.045 & 0.864 & 4.767 & 8.352 \\
\midrule
Fluent & 5.658 & 0.344 & 0.745 & 5.909 & 0.505 & 0.663 & 5.512 & 0.051 & 0.863 & \textbf{5.980} & \textbf{8.636} \\
Simple Language & 5.623 & 0.330 & 0.745 & 5.610 & 0.454 & 0.689 & 5.496 & 0.049 & \textbf{0.873} & 5.196 & 8.412 \\
Technical Terms & 5.618 & 0.327 & 0.771 & 6.142 & 0.548 & 0.681 & 5.566 & 0.057 & 0.875 & 4.411 & 8.582 \\
Authoritative & 5.655 & 0.335 & 0.769 & 5.742 & 0.487 & 0.679 & 5.512 & 0.051 & 0.878 & 4.726 & 7.332 \\
More Quotes & \textbf{5.769} & \textbf{0.365} & 0.731 & \textbf{6.328} & \textbf{0.562} & 0.699 & \textbf{5.642} & 0.081 & 0.872 & 5.077 & 8.440 \\
Citing Credible Sources & 5.689 & 0.341 & 0.733 & 5.908 & 0.489 & 0.696 & 5.572 & 0.062 & 0.867 & 4.624 & 8.153 \\
Statistics & 5.558 & 0.317 & 0.768 & 6.158 & 0.519 & \textbf{0.706} & 5.631 & \textbf{0.091} & 0.862 & 4.705 & 7.613 \\
SEO & 5.640 & 0.333 & 0.765 & 6.138 & 0.534 & 0.691 & 5.576 & 0.057 & 0.881 & 4.817 & 8.419 \\
Unique Words & 5.577 & 0.322 & 0.751 & 5.905 & 0.511 & 0.672 & 5.524 & 0.050 & 0.868 & 4.223 & 8.415 \\
\bottomrule
\end{tabular}
}
\end{table*}

\section{Experiments}
\label{sec:experiments}

In this section, we conduct a comprehensive empirical evaluation of document-level optimization strategies for Generative Search Engines (GSEs). Our analysis aims to answer several key research questions: First, we quantify the overall effectiveness of different optimization heuristics on key performance dimensions including Exposure, Faithful Credit, and Causal Impact (\textbf{RQ1}), and investigate the inherent trade-offs between these objectives and document quality metrics such as Readability and Trustworthiness (\textbf{RQ2}). Furthermore, we examine whether the optimal strategy is context-dependent regarding user intent and query difficulty (\textbf{RQ3}), and how retrieval rank influences the efficacy of these optimizations (\textbf{RQ4}). Finally, we interpret these results by analyzing the linguistic feature shifts associated with successful optimizations (\textbf{RQ5}).

\subsection{Experimental Setup}
\label{sec:setup}

\paragraph{Dataset and Model.}
Our experiments utilize the test split of the \texttt{CC-GSEO-Bench} ($N=5353$), which provides a diverse set of queries accompanied by retrieved documents, designated target documents, and rich metadata tags (e.g., User Intent, Answer Type). Unless otherwise noted, we employ \texttt{gpt-oss-120b} as the backbone GSE model. Results for other architectures and cross-model consistency are provided in Appendix~\ref{sec:appendix_full_results}.

\paragraph{Optimization Strategies.}
We evaluate a suite of nine document rewriting strategies designed to mimic common GSEO heuristics. These include stylistic changes (\textit{Fluent}, \textit{Simple Language}, \textit{Unique Words}), content enrichment (\textit{More Quotes}, \textit{Statistics}, \textit{Citing Credible Sources}, \textit{Technical Terms}), and authority-focused edits (\textit{Authoritative}, \textit{SEO}). A baseline condition, \textit{None}, retains the original target document content. Detailed descriptions of each optimization strategy are provided in Appendix~\ref{sec:appendix_baseline_methods}.

\paragraph{Evaluation Metrics.}
We adopt a multi-dimensional evaluation framework encompassing five core aspects. For the primary influence metrics, \textbf{Exposure (E)}, \textbf{Faithful Credit (F)}, and \textbf{Causal Impact (C)}—we report three aggregated system-level indicators following the GEO metric standard: Mean Influence Level (MIL), Coverage (ICov), and Stability (IStab). Additionally, we monitor document quality through \textbf{Readability (R)} and \textbf{Trustworthiness (T)}, reporting their mean scores. All base metrics are normalized to a 0--10 scale, where higher values indicate better performance.

\begin{figure}[t]
\centering
\includegraphics[width=0.92\linewidth]{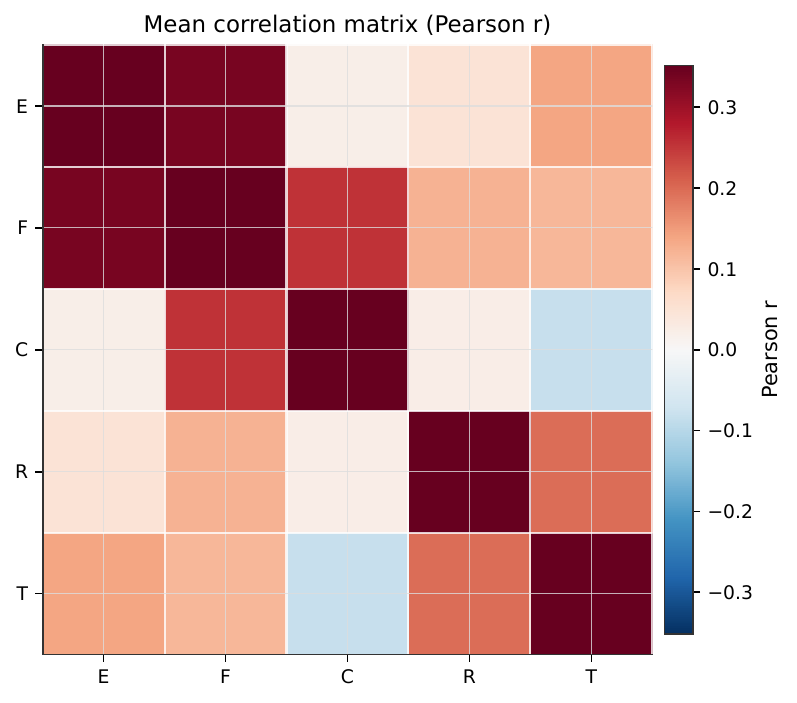}
\caption{Metric correlation heatmap on \texttt{gpt-oss-120b}. Each cell reports the mean Pearson correlation ($r$) between two metric dimensions (E/F/C/R/T), computed at the item level and averaged over optimization runs. The matrix highlights generally weak cross-metric dependencies (e.g., near-zero E--C), alongside several moderate couplings (E--F, F--C, and R--T), supporting the need for multi-objective trade-off analysis.}
\label{fig:rq2_metric_corr}
\vspace{-3mm}
\end{figure}

\subsection{Overall Effectiveness and Trade-offs (RQ1 \& RQ2)}
\label{sec:rq1_rq2}

Table~\ref{tab:main_results} summarizes the system-level performance across all strategies. We observe distinct effectiveness profiles among the optimization methods. Notably, \textit{More Quotes} emerges as the most robust strategy, achieving the highest scores across Exposure (MIL 5.769), Faithful Credit (MIL 6.328), and Causal Impact (MIL 5.642). This suggests that direct quotation facilitates the GSE's ability to attend to and accurately attribute information from the target document.

However, the results also highlight critical trade-offs. While \textit{Statistics} yields substantial gains in Causal Impact (ICov 0.091) and Faithful Credit, it suffers from a significant degradation in Trustworthiness (7.613 vs. 8.352 for baseline), indicating that the aggressive injection of numerical data may be perceived as hallucination-prone or artificially dense. Conversely, \textit{Fluent} rewriting significantly boosts Readability and Trustworthiness but provides only marginal gains in causal influence.

Correlation analysis (Figure~\ref{fig:rq2_metric_corr}) further reveals that the relationship between Exposure (E) and Causal Impact (C) is negligible ($r \approx 0.019$), whereas Faithful Credit (F) shows moderate correlation with C ($r \approx 0.254$). Meanwhile, Readability (R) and Trustworthiness (T) are positively correlated ($r \approx 0.197$), while C and T exhibit a weak negative correlation ($r \approx -0.081$), suggesting a mild influence--quality tension. Overall, the mostly low off-diagonal correlations indicate that these metrics capture complementary aspects rather than being redundant. This implies that mere retrieval visibility is insufficient; the model must be induced to explicitly attribute the source to drive causal changes in the answer. Consequently, \textit{More Quotes}, \textit{Fluent}, and \textit{Technical Terms} constitute the Pareto-optimal frontier when considering the full spectrum of E/F/C/R/T objectives.

\subsection{Contextual and Positional Analysis (RQ3 \& RQ4)}
\label{sec:context_position}

We further investigate the heterogeneity of optimal strategies across diverse user intents and query complexities. Table~\ref{tab:context_best} reports the strategy maximizing Causal Impact ($\Delta$C) for varying query tags. The \textit{More Quotes} strategy demonstrates broad generalization, proving optimal for both ``Learning'' and ``Research'' intents, as well as distinct difficulty levels. Interestingly, for ``Complex'' queries and ``Fact''-seeking questions, \textit{Statistics} outperforms other methods, suggesting that quantitative density becomes a stronger signal for influence when the model requires precise data points.

\begin{table}[h]
\centering
\caption{Optimal strategies maximizing Causal Impact ($\Delta$C) across different query contexts (Intent, Type, Difficulty).}
\label{tab:context_best}
\resizebox{\linewidth}{!}{
\begin{tabular}{llccc}
\toprule
\textbf{Context Tag} & \textbf{Value} & \textbf{Best Method} & $\Delta$\textbf{C} & $\Delta$\textbf{F} \\
\midrule
\multirow{2}{*}{Intent} & Learning & More Quotes & +0.154 & +0.639 \\
 & Research & More Quotes & +0.146 & +0.528 \\
\midrule
\multirow{3}{*}{Type} & Explanation & More Quotes & +0.147 & +0.601 \\
 & Fact & Statistics & +0.115 & +0.259 \\
 & List & More Quotes & +0.233 & +0.523 \\
\midrule
\multirow{3}{*}{Difficulty} & Simple & More Quotes & +0.123 & +0.455 \\
 & Intermediate & More Quotes & +0.159 & +0.612 \\
 & Complex & Statistics & +0.163 & +0.347 \\
\bottomrule
\end{tabular}
}
\vspace{-3mm}
\end{table}

\begin{figure*}[t]
\centering
\includegraphics[width=0.85\linewidth]{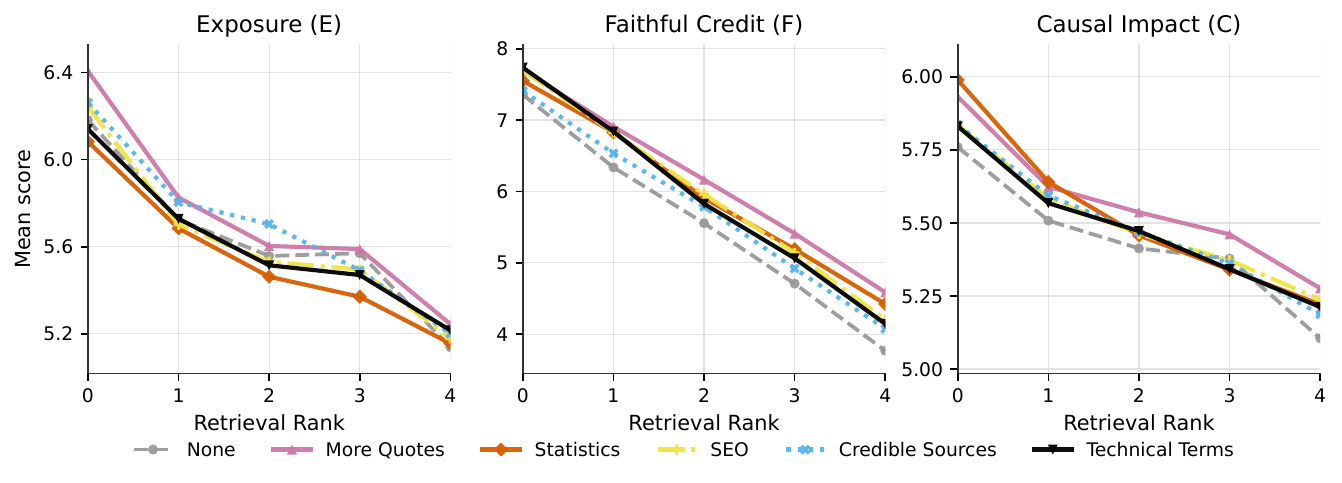}
\caption{Position effect on \texttt{gpt-oss-120b}. We report the mean Exposure (E), Faithful Credit (F), and Causal Impact (C) across retrieval ranks for the \textit{None} baseline and several high-performing optimization strategies; \textit{More Quotes} remains comparatively robust at lower ranks, especially on F and C.}
\label{fig:rq4_position}
\vspace{-4mm}
\end{figure*}

Beyond semantic context, the retrieval position of the target document significantly impacts its influence. As shown in Table~\ref{tab:position_effect}, the baseline influence drops precipitously as the document moves from rank 0 to 4 (F-score declines from 7.36 to 3.77). The \textit{More Quotes} strategy mitigates this position bias, maintaining higher Faithfulness (4.59 vs. 3.77) and Causal Impact at lower ranks. This indicates that optimized content can partially compensate for lower retrieval visibility.
Figure~\ref{fig:rq4_position} compares multiple strategies across positions on Exposure/Faithful Credit/Causal Impact and highlights this resilience trend. This indicates that optimized content can partially compensate for lower retrieval visibility.

\begin{table}[h]
\centering
\caption{Impact of retrieval rank on document influence. \textit{More Quotes} demonstrates greater resilience to position degradation compared to the \textit{None} baseline.}
\label{tab:position_effect}
\resizebox{\linewidth}{!}{
\begin{tabular}{ccccccc}
\toprule
\multirow{2}{*}{\textbf{Pos.}} & \multicolumn{3}{c}{\textbf{None (Baseline)}} & \multicolumn{3}{c}{\textbf{More Quotes}} \\
\cmidrule(lr){2-4} \cmidrule(lr){5-7}
 & E & F & C & E & F & C \\
\midrule
0 & 6.18 & 7.36 & 5.76 & 6.41 & 7.70 & 5.93 \\
1 & 5.73 & 6.34 & 5.51 & 5.83 & 6.91 & 5.62 \\
2 & 5.56 & 5.56 & 5.41 & 5.60 & 6.17 & 5.54 \\
3 & 5.57 & 4.71 & 5.38 & 5.59 & 5.41 & 5.46 \\
4 & 5.14 & 3.77 & 5.11 & 5.25 & 4.59 & 5.28 \\
\bottomrule
\end{tabular}
}
\vspace{-3mm}
\end{table}

\subsection{Mechanism Analysis (RQ5)}
\label{sec:mechanisms}

Finally, to understand the mechanisms driving these improvements, we analyze the linguistic feature shifts induced by each strategy (Table~\ref{tab:features}). As expected, \textit{More Quotes} significantly increases the frequency of quotation markers (+9.7), while \textit{Statistics} introduces a high density of numerical tokens (+7.6). Strategies like \textit{Citing Credible Sources} and \textit{SEO} tend to increase document length (approx. +300-460 characters). A granular correlation analysis reveals that these surface-level edits primarily correlate with improvements in Readability and Trustworthiness rather than directly with Causal Impact, reinforcing the finding that structural formatting (e.g., quotes) serves as a necessary scaffold for influence, rather than a direct causal lever.

\begin{table}[h]
\centering
\caption{Mean change in document features (Optimized - Original). $\Delta$Nums denotes change in numeric tokens excluding citation indices.}
\label{tab:features}
\resizebox{\linewidth}{!}{
\begin{tabular}{lcccc}
\toprule
\textbf{Method} & $\Delta$\textbf{Chars} & $\Delta$\textbf{Words} & $\Delta$\textbf{Nums} & $\Delta$\textbf{Quotes} \\
\midrule
Fluent & -41.2 & -7.9 & -0.2 & +0.2 \\
Simple Lang. & -85.8 & -11.4 & -0.1 & +0.0 \\
Tech. Terms & +100.4 & +1.0 & -0.1 & +0.0 \\
Authoritative & +112.8 & +12.9 & -0.1 & +0.0 \\
More Quotes & +487.6 & +66.7 & +2.5 & \textbf{+9.7} \\
Citing Sources & +462.6 & +66.1 & +2.5 & +0.4 \\
Statistics & +326.5 & +45.9 & \textbf{+7.6} & +0.1 \\
SEO & +303.5 & +39.1 & +0.0 & +0.0 \\
Unique Words & +65.3 & -0.7 & -0.4 & +0.1 \\
\bottomrule
\end{tabular}
}
\vspace{-3mm}
\end{table}

\section{Related Works}

\subsection{Retrieval-Augmented Generation}

Large Language Models (LLMs) often suffer from factual inaccuracies and outdated knowledge~\citep{huang2025survey}. Retrieval-Augmented Generation (RAG) addresses this by connecting LLMs to external, up-to-date knowledge sources~\citep{lewis2020retrieval,gao2023retrieval}. This retrieve-then-generate process first retrieves relevant information, then uses it to enhance the LLM's prompt, leading to more reliable, timely, and trustworthy outputs. Recent advancements include adaptive retrieval~\citep{jiang2023active,liu2024ctrla}, self-corrective mechanisms~\citep{asai2023self} and advanced reasoning strategies~\citep{singh2025agentic}. The field also focuses on robust evaluation frameworks~\citep{yoranmaking} and privacy-preserving techniques~\citep{zeng2024good}.
However, previous work on RAG has typically focused on improving the accuracy of the output~\citep{yu2024evaluation,chen2024benchmarking}, often neglecting the impact of the retrieved content on the final answer~\citep{aggarwal2024geo,wan2024evidence}.

\subsection{Generative Search Engine Optimization}

Generative Search Engines (GSEs), like perplexity, are transforming how we search. Unlike traditional search engines that offer ranked links, GSEs use RAG to synthesize direct, comprehensive, and cited answers from web sources. 
This shift presents a new challenge for content creators, who now need their content to be included and favorably represented within AI-generated responses, rather than just ranking high. This has led to the emergence of Generative Search Engine Optimization (GSEO)~\citep{aggarwal2024geo}. Research indicates that GSEO strategies differ significantly from traditional SEO; instead of keywords, GSEO prioritizes semantic clarity, authoritativeness, and structured data that's easily parsable by an LLM~\citep{aggarwal2024geo,puerto2025c}. 
Because LLMs are very sensitive to the input context~\citep{anagnostidis2024susceptible,wan2024evidence}, jailbreaking has also become a research perspective for GSEO~\citep{pfrommer2024ranking,nestaas2024adversarial}.
Yet, these works overlook the essence of GSEO: it's about boosting content influence across multiple queries, not just one.
\section{Conclusion}

This paper introduces CC-GSEO-Bench and a content-centric evaluation framework for measuring and optimizing source influence in Generative Search Engines. Going beyond surface attribution, our framework quantifies a source's Exposure, Faithful Credit, and Causal Impact, together with content-quality dimensions of Readability \& Structure and Trustworthiness \& Safety. CC-GSEO-Bench pairs each source article with a query cluster and supports article-level influence analysis via strength, coverage, and stability over clustered queries. Experiments on nine representative rewriting strategies reveal systematic trade-offs and strong rank effects, offering actionable guidance for creators and a principled foundation for future GSEO research.

\section*{Limitations}
Our study is conducted in a controlled offline setting with cached retrieval contexts, so absolute results may vary under different deployed GSE pipelines, prompts, or citation behaviors. We rely on automated judges for scalability, which may not capture all subjective or domain-specific notions of readability and trust without complementary human evaluation. Finally, we focus on document-level rewriting and do not model broader ecosystem factors such as personalization, reputation signals, or temporal drift, which are important directions for future work.

\section*{Ethical Considerations}
The ethical implications of this work require careful consideration. A primary concern is that these GSEO techniques could be misused to amplify misinformation or spam, leading to an arms race that prioritizes machine manipulation over human value and degrades web quality. Our system could also learn deceptive tactics, such as inventing citations, to artificially boost influence. Finally, unequal access to such powerful optimization tools could allow well-resourced groups to dominate search results, centralizing influence and reducing the diversity of voices online.
\bibliography{custom}

\begin{thebibliography}{31}
\providecommand{\natexlab}[1]{#1}

\bibitem[{Achiam et~al.(2023)Achiam, Adler, Agarwal, Ahmad, Akkaya, Aleman, Almeida, Altenschmidt, Altman, Anadkat et~al.}]{achiam2023gpt}
Josh Achiam, Steven Adler, Sandhini Agarwal, Lama Ahmad, Ilge Akkaya, Florencia~Leoni Aleman, Diogo Almeida, Janko Altenschmidt, Sam Altman, Shyamal Anadkat, et~al. 2023.
\newblock Gpt-4 technical report.
\newblock \emph{arXiv preprint arXiv:2303.08774}.

\bibitem[{Aggarwal et~al.(2024)Aggarwal, Murahari, Rajpurohit, Kalyan, Narasimhan, and Deshpande}]{aggarwal2024geo}
Pranjal Aggarwal, Vishvak Murahari, Tanmay Rajpurohit, Ashwin Kalyan, Karthik Narasimhan, and Ameet Deshpande. 2024.
\newblock Geo: Generative engine optimization.
\newblock In \emph{Proceedings of the 30th ACM SIGKDD Conference on Knowledge Discovery and Data Mining}, pages 5--16.

\bibitem[{Allan et~al.(2024)Allan, Choi, Lopresti, and Zamani}]{allan2024future}
James Allan, Eunsol Choi, Daniel~P Lopresti, and Hamed Zamani. 2024.
\newblock Future of information retrieval research in the age of generative ai.
\newblock \emph{arXiv preprint arXiv:2412.02043}.

\bibitem[{Anagnostidis and Bulian(2024)}]{anagnostidis2024susceptible}
Sotiris Anagnostidis and Jannis Bulian. 2024.
\newblock How susceptible are llms to influence in prompts?
\newblock \emph{arXiv preprint arXiv:2408.11865}.

\bibitem[{Arasu et~al.(2001)Arasu, Cho, Garcia-Molina, Paepcke, and Raghavan}]{arasu2001searching}
Arvind Arasu, Junghoo Cho, Hector Garcia-Molina, Andreas Paepcke, and Sriram Raghavan. 2001.
\newblock Searching the web.
\newblock \emph{ACM Transactions on Internet Technology (TOIT)}, 1(1):2--43.

\bibitem[{Asai et~al.(2023)Asai, Wu, Wang, Sil, and Hajishirzi}]{asai2023self}
Akari Asai, Zeqiu Wu, Yizhong Wang, Avirup Sil, and Hannaneh Hajishirzi. 2023.
\newblock Self-rag: Learning to retrieve, generate, and critique through self-reflection.
\newblock In \emph{The Twelfth International Conference on Learning Representations}.

\bibitem[{Chen et~al.(2024)Chen, Lin, Han, and Sun}]{chen2024benchmarking}
Jiawei Chen, Hongyu Lin, Xianpei Han, and Le~Sun. 2024.
\newblock Benchmarking large language models in retrieval-augmented generation.
\newblock In \emph{Proceedings of the AAAI Conference on Artificial Intelligence}, volume~38, pages 17754--17762.

\bibitem[{Davis(2006)}]{davis2006search}
Harold Davis. 2006.
\newblock \emph{Search engine optimization}.
\newblock " O'Reilly Media, Inc.".

\bibitem[{Fan et~al.(2019)Fan, Jernite, Perez, Grangier, Weston, and Auli}]{fan2019eli5}
Angela Fan, Yacine Jernite, Ethan Perez, David Grangier, Jason Weston, and Michael Auli. 2019.
\newblock Eli5: Long form question answering.
\newblock In \emph{Proceedings of the 57th Annual Meeting of the Association for Computational Linguistics}, pages 3558--3567.

\bibitem[{Gao et~al.(2023)Gao, Xiong, Gao, Jia, Pan, Bi, Dai, Sun, Wang, and Wang}]{gao2023retrieval}
Yunfan Gao, Yun Xiong, Xinyu Gao, Kangxiang Jia, Jinliu Pan, Yuxi Bi, Yixin Dai, Jiawei Sun, Haofen Wang, and Haofen Wang. 2023.
\newblock Retrieval-augmented generation for large language models: A survey.
\newblock \emph{arXiv preprint arXiv:2312.10997}, 2(1).

\bibitem[{Hu et~al.(2024)Hu, Chen, Li, Guo, Wen, Yu, and Guo}]{hu2024towards}
Xuming Hu, Junzhe Chen, Xiaochuan Li, Yufei Guo, Lijie Wen, Philip~S Yu, and Zhijiang Guo. 2024.
\newblock Towards understanding factual knowledge of large language models.
\newblock In \emph{The twelfth international conference on learning representations}.

\bibitem[{Huang et~al.(2025)Huang, Yu, Ma, Zhong, Feng, Wang, Chen, Peng, Feng, Qin et~al.}]{huang2025survey}
Lei Huang, Weijiang Yu, Weitao Ma, Weihong Zhong, Zhangyin Feng, Haotian Wang, Qianglong Chen, Weihua Peng, Xiaocheng Feng, Bing Qin, et~al. 2025.
\newblock A survey on hallucination in large language models: Principles, taxonomy, challenges, and open questions.
\newblock \emph{ACM Transactions on Information Systems}, 43(2):1--55.

\bibitem[{Jiang et~al.(2023)Jiang, Xu, Gao, Sun, Liu, Dwivedi-Yu, Yang, Callan, and Neubig}]{jiang2023active}
Zhengbao Jiang, Frank~F Xu, Luyu Gao, Zhiqing Sun, Qian Liu, Jane Dwivedi-Yu, Yiming Yang, Jamie Callan, and Graham Neubig. 2023.
\newblock Active retrieval augmented generation.
\newblock In \emph{Proceedings of the 2023 Conference on Empirical Methods in Natural Language Processing}, pages 7969--7992.

\bibitem[{Kwiatkowski et~al.(2019)Kwiatkowski, Palomaki, Redfield, Collins, Parikh, Alberti, Epstein, Polosukhin, Devlin, Lee et~al.}]{kwiatkowski2019natural}
Tom Kwiatkowski, Jennimaria Palomaki, Olivia Redfield, Michael Collins, Ankur Parikh, Chris Alberti, Danielle Epstein, Illia Polosukhin, Jacob Devlin, Kenton Lee, et~al. 2019.
\newblock Natural questions: a benchmark for question answering research.
\newblock \emph{Transactions of the Association for Computational Linguistics}, 7:453--466.

\bibitem[{Lewis et~al.(2020)Lewis, Perez, Piktus, Petroni, Karpukhin, Goyal, K{\"u}ttler, Lewis, Yih, Rockt{\"a}schel et~al.}]{lewis2020retrieval}
Patrick Lewis, Ethan Perez, Aleksandra Piktus, Fabio Petroni, Vladimir Karpukhin, Naman Goyal, Heinrich K{\"u}ttler, Mike Lewis, Wen-tau Yih, Tim Rockt{\"a}schel, et~al. 2020.
\newblock Retrieval-augmented generation for knowledge-intensive nlp tasks.
\newblock \emph{Advances in neural information processing systems}, 33:9459--9474.

\bibitem[{Liu et~al.(2024)Liu, Zhang, Guo, Dong, Li, Lee, Zhang, and Liu}]{liu2024ctrla}
Huanshuo Liu, Hao Zhang, Zhijiang Guo, Kuicai Dong, Xiangyang Li, Yi~Quan Lee, Cong Zhang, and Yong Liu. 2024.
\newblock Ctrla: Adaptive retrieval-augmented generation via probe-guided control.
\newblock \emph{arXiv e-prints}, pages arXiv--2405.

\bibitem[{Liu et~al.(2023)Liu, Zhang, and Liang}]{liu2023evaluating}
Nelson~F Liu, Tianyi Zhang, and Percy Liang. 2023.
\newblock Evaluating verifiability in generative search engines.
\newblock \emph{arXiv preprint arXiv:2304.09848}.

\bibitem[{Nestaas et~al.(2024)Nestaas, Debenedetti, and Tram{\`e}r}]{nestaas2024adversarial}
Fredrik Nestaas, Edoardo Debenedetti, and Florian Tram{\`e}r. 2024.
\newblock Adversarial search engine optimization for large language models.
\newblock \emph{arXiv preprint arXiv:2406.18382}.

\bibitem[{Nguyen et~al.(2016)Nguyen, Rosenberg, Song, Gao, Tiwary, Majumder, and Deng}]{nguyen2016ms}
Tri Nguyen, Mir Rosenberg, Xia Song, Jianfeng Gao, Saurabh Tiwary, Rangan Majumder, and Li~Deng. 2016.
\newblock Ms marco: A human-generated machine reading comprehension dataset.

\bibitem[{Ouyang et~al.(2022)Ouyang, Wu, Jiang, Almeida, Wainwright, Mishkin, Zhang, Agarwal, Slama, Ray et~al.}]{ouyang2022training}
Long Ouyang, Jeffrey Wu, Xu~Jiang, Diogo Almeida, Carroll Wainwright, Pamela Mishkin, Chong Zhang, Sandhini Agarwal, Katarina Slama, Alex Ray, et~al. 2022.
\newblock Training language models to follow instructions with human feedback.
\newblock \emph{Advances in neural information processing systems}, 35:27730--27744.

\bibitem[{Pfrommer et~al.(2024)Pfrommer, Bai, Gautam, and Sojoudi}]{pfrommer2024ranking}
Samuel Pfrommer, Yatong Bai, Tanmay Gautam, and Somayeh Sojoudi. 2024.
\newblock Ranking manipulation for conversational search engines.
\newblock \emph{arXiv preprint arXiv:2406.03589}.

\bibitem[{Puerto et~al.(2025)Puerto, Gubri, Green, Oh, and Yun}]{puerto2025c}
Haritz Puerto, Martin Gubri, Tommaso Green, Seong~Joon Oh, and Sangdoo Yun. 2025.
\newblock C-seo bench: Does conversational seo work?
\newblock \emph{arXiv preprint arXiv:2506.11097}.

\bibitem[{Sch{\"u}tze et~al.(2008)Sch{\"u}tze, Manning, and Raghavan}]{schutze2008introduction}
Hinrich Sch{\"u}tze, Christopher~D Manning, and Prabhakar Raghavan. 2008.
\newblock \emph{Introduction to information retrieval}, volume~39.
\newblock Cambridge University Press Cambridge.

\bibitem[{Schwartz(1998)}]{schwartz1998web}
Candy Schwartz. 1998.
\newblock Web search engines.
\newblock \emph{Journal of the American Society for Information Science}, 49(11):973--982.

\bibitem[{Singh et~al.(2025)Singh, Ehtesham, Kumar, and Khoei}]{singh2025agentic}
Aditi Singh, Abul Ehtesham, Saket Kumar, and Tala~Talaei Khoei. 2025.
\newblock Agentic retrieval-augmented generation: A survey on agentic rag.
\newblock \emph{arXiv preprint arXiv:2501.09136}.

\bibitem[{Wan et~al.(2024)Wan, Wallace, and Klein}]{wan2024evidence}
Alexander Wan, Eric Wallace, and Dan Klein. 2024.
\newblock What evidence do language models find convincing?
\newblock \emph{arXiv preprint arXiv:2402.11782}.

\bibitem[{Xu et~al.(2024)Xu, Qi, Qi, Xu, and Guo}]{xu2024debateqa}
Rongwu Xu, Xuan Qi, Zehan Qi, Wei Xu, and Zhijiang Guo. 2024.
\newblock Debateqa: Evaluating question answering on debatable knowledge.
\newblock \emph{arXiv preprint arXiv:2408.01419}.

\bibitem[{Yang et~al.(2018)Yang, Qi, Zhang, Bengio, Cohen, Salakhutdinov, and Manning}]{yang2018hotpotqa}
Zhilin Yang, Peng Qi, Saizheng Zhang, Yoshua Bengio, William~W Cohen, Ruslan Salakhutdinov, and Christopher~D Manning. 2018.
\newblock Hotpotqa: A dataset for diverse, explainable multi-hop question answering.
\newblock \emph{arXiv preprint arXiv:1809.09600}.

\bibitem[{Yoran et~al.()Yoran, Wolfson, Ram, and Berant}]{yoranmaking}
Ori Yoran, Tomer Wolfson, Ori Ram, and Jonathan Berant.
\newblock Making retrieval-augmented language models robust to irrelevant context.
\newblock In \emph{The Twelfth International Conference on Learning Representations}.

\bibitem[{Yu et~al.(2024)Yu, Gan, Zhang, Tong, Liu, and Liu}]{yu2024evaluation}
Hao Yu, Aoran Gan, Kai Zhang, Shiwei Tong, Qi~Liu, and Zhaofeng Liu. 2024.
\newblock Evaluation of retrieval-augmented generation: A survey.
\newblock In \emph{CCF Conference on Big Data}, pages 102--120. Springer.

\bibitem[{Zeng et~al.(2024)Zeng, Zhang, He, Xing, Liu, Xu, Ren, Wang, Yin, Chang et~al.}]{zeng2024good}
Shenglai Zeng, Jiankun Zhang, Pengfei He, Yue Xing, Yiding Liu, Han Xu, Jie Ren, Shuaiqiang Wang, Dawei Yin, Yi~Chang, et~al. 2024.
\newblock The good and the bad: Exploring privacy issues in retrieval-augmented generation (rag).
\newblock \emph{arXiv preprint arXiv:2402.16893}.

\end{thebibliography}

\appendix
\newpage
\section{Baseline GSEO Methods}
\label{sec:appendix_baseline_methods}

Our approach leverages distinct prompting strategies to achieve various GSEO goals, each targeting a specific aspect of content presentation to improve its ``rank'' within a LLMs generated response. These baselines aim to enhance a source's visibility within answers generated by Generative Search Engines (GSEs), ultimately increasing the likelihood of citation.

\subsection{Textual Fluency and Engagement}

\begin{itemize}
    \item \textbf{Fluent}: This method focuses on refining the prose to be more natural and engaging. By rephrasing sentences for better flow and clarity, the goal is to make the content more appealing and digestible for the LLM, potentially increasing its selection for inclusion.
    \item \textbf{Simple Language}: This strategy prioritizes clarity and ease of understanding. It simplifies the language to ensure the core information is conveyed as directly as possible, potentially making the source more broadly accessible and thus more frequently cited.
    \item \textbf{Technical Terms}: This method introduces more technical terms and factual language, aiming to present the existing information in a more specialized and authoritative manner. This could make the source more appealing for technical queries or when a more in-depth explanation is required.
\end{itemize}

\subsection{Authority and Credibility Building}

\begin{itemize}
    \item \textbf{Authoritative}: This method seeks to imbue the source text with a confident and expert tone. By using assertive language and phrases that convey strong guarantees or unique value, the intent is to signal the source's definitive nature and increase its perceived authority.
    \item \textbf{More Quotes}: This baseline focuses on integrating additional quotes into the text. These quotes, even if artificial, are designed to appear reputable, thereby enhancing the perceived influence and importance of the source material. The underlying idea is that content backed by ``external'' validation might be favored.
    \item \textbf{Citing Sources}: This strategy involves naturally incorporating plausible citations to credible (though potentially invented) sources. The aim is to make the original source appear well-researched and attended to by experts, thus boosting its perceived reliability and trustworthiness.
    \item \textbf{Statistics}: This method strategically injects positive and compelling statistics throughout the text. By adding objective numerical facts, even hypothetical ones, the goal is to enhance the source's credibility and make its claims more concrete and persuasive to the GSEs.
\end{itemize}

\subsection{SEO Techniques}

\begin{itemize}
    \item \textbf{Unique Words}: This baseline aims to enrich the vocabulary of the source text by incorporating more unique and less common words. The hypothesis here is that a richer, more diverse vocabulary might signal higher quality or more specialized content, potentially making it more attractive for an LLM to select and cite. This can be seen as a form of ``vocabulary stuffing'' for generative models, distinct from traditional keyword stuffing.
    \item \textbf{SEO}: This method directly addresses traditional SEO principles by incorporating new, relevant keywords into the source text. The objective is to make the content more discoverable and relevant to a wider range of queries, anticipating that GSEs will still consider keyword relevance in their answer generation process. This is a direct application of what might be termed ``keyword stuffing'' specifically for generative search, focusing on explicit keyword integration.
\end{itemize}

These diverse baseline methods provide a robust framework for evaluating how different textual manipulations, guided by specific GSEO principles, can influence the visibility and citation frequency of web sources within the context of generative search. By comparing the effectiveness of these approaches, we can gain valuable insights into the optimal strategies for GSEO.

\section{Benchmark Construction}
\label{benchmark_construction}

\subsection{Construction Process}

The construction of the CC-GSEO-Bench followed a systematic, multi-stage process designed to ensure content-centricity and ecological validity. The process commenced with the aggregation of a diverse set of seed queries from a wide array of publicly available, open-source datasets. This collection included queries from established benchmarks such as Pinocchio~\citep{hu2024towards}, ELI5 (Explain Like I'm 5)~\citep{fan2019eli5}, Natural Questions (NQ)~\citep{kwiatkowski2019natural}, HotpotQA~\citep{yang2018hotpotqa}, DebateQA~\citep{xu2024debateqa}, MS MARCO~\citep{nguyen2016ms}, and AllSouls~\citep{liu2023evaluating} to ensure broad topical coverage and a rich variety of query formulations.

Each of these seed queries was then programmatically submitted to the Tavily search API to retrieve a set of relevant documents. From the ranked list of results returned for each query, we retained the top 10 documents with the highest relevance scores. Subsequently, from this candidate pool, one document was randomly selected to serve as a source article. 

With a corpus of source articles established, the next stage was to generate a set of user queries tailored to each article. For every source article, we utilized the \texttt{gpt-5} model to simulate realistic user queries that could be answered by the article's content. This procedure shifts the paradigm from a traditional query-centric view to a content-centric one, where the article itself dictates the scope of relevant inquiries.

To ensure a strong, verifiable link between the simulated queries and their corresponding source articles, a final filtering stage was implemented. Each newly generated query was used to execute a new search operation with the Tavily API. A query was only retained in the final benchmark if its corresponding source article appeared within the new set of search results. This crucial verification step guarantees that the query-article pairs in CC-GSEO-Bench are not just semantically related but are also contextually linked within a realistic retrieval framework.

\subsection{Benchmark Statistics}

Our data construction process culminates in the CC-GSEO-Bench, a substantial benchmark designed for content-centric evaluation. The benchmark consists of \textbf{1,030 unique source articles}, each paired with a set of relevant user inquiries, totaling \textbf{5,353 query-article pairs}. The statistical distributions underpinning the benchmark are provided in Tables \ref{tab:source-dist}, \ref{tab:query-dist}, and \ref{tab:rank-dist}. These statistics underscore the diversity and scale of the dataset.

Table \ref{tab:source-dist} details the breakdown of the query-article pairs according to the original dataset from which the seed queries were sourced. Table \ref{tab:query-dist} illustrates the one-to-many relationship central to our content-centric design, showing the distribution of the number of queries associated with each article. Finally, Table \ref{tab:rank-dist} presents the distribution of the ranks of our source documents within the search engine results, which confirms the high relevance of the selected articles to the generated queries.

\begin{table}[h!]
\centering
\caption{Distribution of Query-Article Pairs by Original Source Dataset.}
\label{tab:source-dist}
\begin{tabular}{lc}
\toprule
\textbf{Original Source Dataset} & \textbf{Number of Pairs} \\
\midrule
ELI5 & 1710 \\
Pinocchio & 1017 \\
Natural Questions (NQ) & 922 \\
MS MARCO & 920 \\
HotpotQA & 197 \\
DebateQA & 185 \\
AllSouls & 49 \\
\bottomrule
\end{tabular}
\vspace{-4mm}
\end{table}

\begin{table}[h!]
\centering
\caption{Distribution of Queries per Source Article.}
\label{tab:query-dist}
\begin{tabular}{cc}
\toprule
\textbf{Queries per Article} & \textbf{Number of Articles} \\
\midrule
2 & 1 \\
3 & 272 \\
4 & 212 \\
5 & 150 \\
6 & 113 \\
7 & 114 \\
8 & 80 \\
9 & 59 \\
10 & 29 \\
\bottomrule
\end{tabular}
\vspace{-4mm}
\end{table}

\begin{table}[h!]
\centering
\caption{Distribution of Source Document Ranks in Search Results.}
\label{tab:rank-dist}
\begin{tabular}{cc}
\toprule
\textbf{Rank in Search Results} & \textbf{Count} \\
\midrule
0 & 1882 \\
1 & 839 \\
2 & 608 \\
3 & 521 \\
4 & 1503 \\
\bottomrule
\end{tabular}
\vspace{-4mm}
\end{table}

\section{Full Results for GPT-5-Nano and GPT-OSS-120B}
\label{sec:appendix_full_results}

This appendix consolidates the full experimental results for \texttt{gpt-5-nano} (``GPT-5-Nano'') and \texttt{gpt-oss-120b} (``GPT-OSS-120B'') on CC-GSEO-Bench.
Unless otherwise specified, all reported scores are averages over 5,353 test instances and are judged on an integer 0--10 scale.
We report five evaluation dimensions: Exposure (E), Faithful Credit (F), Causal Impact (C), Readability \& Structure (R), and Trustworthiness \& Safety (T).

\subsection{Overall Effects of Document Optimization}

Table~\ref{tab:app_overall_means_nano} and Table~\ref{tab:app_overall_means_oss} report the item-level mean scores for each optimization strategy.
Table~\ref{tab:app_overall_deltas_nano} and Table~\ref{tab:app_overall_deltas_oss} additionally report the deltas relative to the unoptimized baseline (None).
Overall, \emph{More Quotes} yields the strongest and most consistent gains on E/F/C, while \emph{Statistics} produces the largest improvement in C but noticeably reduces T.
\emph{Authoritative} and \emph{Fluent} mainly improve R, but do not always improve C.

\begin{table}[h]
  \centering
  \scriptsize
  \setlength{\tabcolsep}{3pt}
  \caption{GPT-5-Nano: Item-level mean scores (0--10) across optimization strategies.}
  \label{tab:app_overall_means_nano}
  \resizebox{\columnwidth}{!}{
    \begin{tabular}{lrrrrr}
      \hline
      Strategy & E & F & C & R & T \\
      \hline
      None & 5.711 & 5.722 & 5.462 & 4.753 & 8.437 \\
      Authoritative & 5.710 & 5.822 & 5.454 & 5.293 & 8.562 \\
      Credible Sources & 5.762 & 5.806 & 5.513 & 4.543 & 8.278 \\
      Fluent & 5.698 & 5.726 & 5.440 & 5.285 & 8.550 \\
      More Quotes & 5.803 & 6.290 & 5.563 & 4.907 & 8.592 \\
      SEO & 5.711 & 5.978 & 5.510 & 4.648 & 8.486 \\
      Simple Language & 5.699 & 5.642 & 5.441 & 5.077 & 8.481 \\
      Statistics & 5.638 & 6.023 & 5.600 & 4.577 & 7.715 \\
      Technical Terms & 5.683 & 5.898 & 5.491 & 4.775 & 8.585 \\
      Unique Words & 5.662 & 5.778 & 5.457 & 4.794 & 8.564 \\
      \hline
    \end{tabular}
  }
\end{table}

\begin{table}[h]
  \centering
  \scriptsize
  \setlength{\tabcolsep}{3pt}
  \caption{GPT-OSS-120B: Item-level mean scores (0--10) across optimization strategies.}
  \label{tab:app_overall_means_oss}
  \resizebox{\columnwidth}{!}{
    \begin{tabular}{lrrrrr}
      \hline
      Strategy & E & F & C & R & T \\
      \hline
      None & 5.686 & 5.727 & 5.460 & 4.750 & 8.439 \\
      Authoritative & 5.726 & 5.727 & 5.469 & 4.725 & 7.400 \\
      Credible Sources & 5.754 & 5.910 & 5.528 & 4.610 & 8.220 \\
      Fluent & 5.722 & 5.873 & 5.483 & 5.929 & 8.700 \\
      More Quotes & 5.819 & 6.304 & 5.608 & 5.063 & 8.490 \\
      SEO & 5.705 & 6.134 & 5.540 & 4.796 & 8.474 \\
      Simple Language & 5.688 & 5.610 & 5.460 & 5.181 & 8.484 \\
      Statistics & 5.620 & 6.143 & 5.595 & 4.671 & 7.699 \\
      Technical Terms & 5.680 & 6.111 & 5.527 & 4.412 & 8.659 \\
      Unique Words & 5.648 & 5.870 & 5.485 & 4.244 & 8.484 \\
      \hline
    \end{tabular}
  }
\end{table}

\begin{table}[t]
  \centering
  \scriptsize
  \setlength{\tabcolsep}{3pt}
  \caption{GPT-5-Nano: Deltas ($\Delta$) relative to the unoptimized baseline (None).}
  \label{tab:app_overall_deltas_nano}
  \resizebox{\columnwidth}{!}{
    \begin{tabular}{lrrrrr}
      \hline
      Strategy & $\Delta$E & $\Delta$F & $\Delta$C & $\Delta$R & $\Delta$T \\
      \hline
      Authoritative & -0.001 & +0.101 & -0.008 & +0.541 & +0.125 \\
      Credible Sources & +0.051 & +0.084 & +0.051 & -0.209 & -0.159 \\
      Fluent & -0.013 & +0.005 & -0.023 & +0.533 & +0.113 \\
      More Quotes & +0.092 & +0.568 & +0.101 & +0.154 & +0.155 \\
      SEO & +0.000 & +0.256 & +0.047 & -0.104 & +0.048 \\
      Simple Language & -0.012 & -0.080 & -0.022 & +0.325 & +0.044 \\
      Statistics & -0.073 & +0.302 & +0.138 & -0.175 & -0.722 \\
      Technical Terms & -0.028 & +0.176 & +0.029 & +0.023 & +0.148 \\
      Unique Words & -0.049 & +0.057 & -0.006 & +0.041 & +0.127 \\
      \hline
    \end{tabular}
  }
\end{table}

\begin{table}[t]
  \centering
  \scriptsize
  \setlength{\tabcolsep}{3pt}
  \caption{GPT-OSS-120B: Deltas ($\Delta$) relative to the unoptimized baseline (None).}
  \label{tab:app_overall_deltas_oss}
  \resizebox{\columnwidth}{!}{
    \begin{tabular}{lrrrrr}
      \hline
      Strategy & $\Delta$E & $\Delta$F & $\Delta$C & $\Delta$R & $\Delta$T \\
      \hline
      Authoritative & +0.040 & +0.000 & +0.010 & -0.025 & -1.038 \\
      Credible Sources & +0.067 & +0.183 & +0.068 & -0.140 & -0.219 \\
      Fluent & +0.035 & +0.146 & +0.023 & +1.179 & +0.261 \\
      More Quotes & +0.132 & +0.577 & +0.149 & +0.313 & +0.052 \\
      SEO & +0.019 & +0.407 & +0.080 & +0.046 & +0.035 \\
      Simple Language & +0.002 & -0.117 & +0.000 & +0.431 & +0.046 \\
      Statistics & -0.067 & +0.416 & +0.135 & -0.079 & -0.739 \\
      Technical Terms & -0.006 & +0.384 & +0.068 & -0.338 & +0.220 \\
      Unique Words & -0.039 & +0.143 & +0.025 & -0.506 & +0.045 \\
      \hline
    \end{tabular}
  }
\end{table}

\subsection{GEO System-Level Metrics (Article-Level Aggregation)}

In addition to item-level means, we report GEO system-level aggregation metrics computed at the article level:
Mean Influence Level (MIL), Influence Coverage (ICov, i.e., the fraction of queries above a threshold), and Influence Stability (IStab, i.e., variance-based stability normalized by a capped max variance).
These GEO metrics are computed for E/F/C, while R/T are summarized by MIL.

\begin{table}[t]
  \centering
  \scriptsize
  \setlength{\tabcolsep}{3pt}
  \caption{GPT-5-Nano: GEO system-level MIL metrics aggregated at the article level.}
  \label{tab:app_geo_mil_nano}
  \resizebox{\columnwidth}{!}{
    \begin{tabular}{lrrrrr}
      \hline
      Strategy & E.MIL & F.MIL & C.MIL & R.MIL & T.MIL \\
      \hline
      None & 5.651 & 5.738 & 5.498 & 4.758 & 8.371 \\
      Authoritative & 5.663 & 5.850 & 5.502 & 5.322 & 8.486 \\
      Credible Sources & 5.701 & 5.821 & 5.546 & 4.496 & 8.203 \\
      Fluent & 5.628 & 5.745 & 5.474 & 5.316 & 8.486 \\
      More Quotes & 5.748 & 6.321 & 5.597 & 4.888 & 8.527 \\
      SEO & 5.649 & 5.994 & 5.551 & 4.707 & 8.426 \\
      Simple Language & 5.639 & 5.665 & 5.486 & 5.124 & 8.402 \\
      Statistics & 5.582 & 6.055 & 5.642 & 4.641 & 7.656 \\
      Technical Terms & 5.624 & 5.936 & 5.534 & 4.823 & 8.523 \\
      Unique Words & 5.607 & 5.799 & 5.498 & 4.822 & 8.489 \\
      \hline
    \end{tabular}
  }
\end{table}

\begin{table}[t]
  \centering
  \scriptsize
  \setlength{\tabcolsep}{3pt}
  \caption{GPT-OSS-120B: GEO system-level MIL metrics aggregated at the article level.}
  \label{tab:app_geo_mil_oss}
  \resizebox{\columnwidth}{!}{
    \begin{tabular}{lrrrrr}
      \hline
      Strategy & E.MIL & F.MIL & C.MIL & R.MIL & T.MIL \\
      \hline
      None & 5.630 & 5.747 & 5.501 & 4.767 & 8.352 \\
      Authoritative & 5.655 & 5.742 & 5.512 & 4.726 & 7.332 \\
      Credible Sources & 5.689 & 5.908 & 5.572 & 4.624 & 8.153 \\
      Fluent & 5.658 & 5.909 & 5.512 & 5.980 & 8.636 \\
      More Quotes & 5.769 & 6.327 & 5.642 & 5.077 & 8.440 \\
      SEO & 5.640 & 6.138 & 5.576 & 4.817 & 8.419 \\
      Simple Language & 5.623 & 5.610 & 5.496 & 5.196 & 8.412 \\
      Statistics & 5.558 & 6.158 & 5.631 & 4.705 & 7.613 \\
      Technical Terms & 5.618 & 6.142 & 5.566 & 4.411 & 8.582 \\
      Unique Words & 5.577 & 5.904 & 5.524 & 4.223 & 8.415 \\
      \hline
    \end{tabular}
  }
\end{table}

\begin{table}[t]
  \centering
  \scriptsize
  \setlength{\tabcolsep}{2pt}
  \caption{GPT-5-Nano: GEO coverage (ICov) and stability (IStab) for E/F/C, aggregated at the article level.}
  \label{tab:app_geo_cov_stab_nano}
  \resizebox{\columnwidth}{!}{
    \begin{tabular}{lrrrrrr}
      \hline
      Strategy & E.ICov & E.IStab & F.ICov & F.IStab & C.ICov & C.IStab \\
      \hline
      None & 0.335 & 0.749 & 0.474 & 0.700 & 0.045 & 0.878 \\
      Authoritative & 0.333 & 0.739 & 0.501 & 0.676 & 0.051 & 0.859 \\
      Credible Sources & 0.344 & 0.736 & 0.486 & 0.681 & 0.059 & 0.866 \\
      Fluent & 0.334 & 0.773 & 0.484 & 0.680 & 0.050 & 0.869 \\
      More Quotes & 0.352 & 0.739 & 0.567 & 0.683 & 0.062 & 0.880 \\
      SEO & 0.333 & 0.743 & 0.514 & 0.704 & 0.057 & 0.875 \\
      Simple Language & 0.331 & 0.729 & 0.465 & 0.689 & 0.049 & 0.856 \\
      Statistics & 0.321 & 0.777 & 0.519 & 0.712 & 0.091 & 0.887 \\
      Technical Terms & 0.328 & 0.715 & 0.509 & 0.684 & 0.050 & 0.876 \\
      Unique Words & 0.328 & 0.774 & 0.494 & 0.671 & 0.048 & 0.876 \\
      \hline
    \end{tabular}
  }
\end{table}

\begin{table}[t]
  \centering
  \scriptsize
  \setlength{\tabcolsep}{2pt}
  \caption{GPT-OSS-120B: GEO coverage (ICov) and stability (IStab) for E/F/C, aggregated at the article level.}
  \label{tab:app_geo_cov_stab_oss}
  \resizebox{\columnwidth}{!}{
    \begin{tabular}{lrrrrrr}
      \hline
      Strategy & E.ICov & E.IStab & F.ICov & F.IStab & C.ICov & C.IStab \\
      \hline
      None & 0.333 & 0.773 & 0.477 & 0.691 & 0.045 & 0.864 \\
      Authoritative & 0.335 & 0.769 & 0.487 & 0.679 & 0.051 & 0.878 \\
      Credible Sources & 0.341 & 0.733 & 0.489 & 0.696 & 0.062 & 0.867 \\
      Fluent & 0.344 & 0.745 & 0.505 & 0.663 & 0.051 & 0.863 \\
      More Quotes & 0.365 & 0.731 & 0.562 & 0.699 & 0.081 & 0.872 \\
      SEO & 0.333 & 0.765 & 0.534 & 0.691 & 0.057 & 0.881 \\
      Simple Language & 0.330 & 0.745 & 0.454 & 0.689 & 0.049 & 0.873 \\
      Statistics & 0.317 & 0.768 & 0.519 & 0.706 & 0.091 & 0.862 \\
      Technical Terms & 0.327 & 0.771 & 0.548 & 0.681 & 0.057 & 0.875 \\
      Unique Words & 0.322 & 0.751 & 0.511 & 0.672 & 0.050 & 0.868 \\
      \hline
    \end{tabular}
  }
\end{table}

\subsection{Metric Trade-offs and Pareto Frontier}

We compute within-run Pearson correlations between metrics and summarize their ranges across all strategies.
E and F are moderately correlated, while E and C are near-uncorrelated; F and C show a mild positive correlation (Table~\ref{tab:app_correlations_nano} and Table~\ref{tab:app_correlations_oss}).
We also report the Pareto-optimal strategies under (i) E/F/C only and (ii) all five metrics (Table~\ref{tab:app_rq2_pareto}).

\begin{table}[t]
  \centering
  \scriptsize
  \setlength{\tabcolsep}{6pt}
  \caption{GPT-5-Nano: Summary statistics of within-run Pearson correlations between metrics (across strategies, including the baseline).}
  \label{tab:app_correlations_nano}
  \begin{tabular}{lrrr}
    \hline
    Pair & Mean & Min & Max \\
    \hline
    E--F & 0.341 & 0.269 & 0.377 \\
    F--C & 0.261 & 0.235 & 0.300 \\
    E--C & 0.022 & 0.007 & 0.033 \\
    R--T & 0.200 & 0.087 & 0.281 \\
    \hline
  \end{tabular}
\end{table}

\begin{table}[t]
  \centering
  \scriptsize
  \setlength{\tabcolsep}{6pt}
  \caption{GPT-OSS-120B: Summary statistics of within-run Pearson correlations between metrics (across strategies, including the baseline).}
  \label{tab:app_correlations_oss}
  \begin{tabular}{lrrr}
    \hline
    Pair & Mean & Min & Max \\
    \hline
    E--F & 0.335 & 0.277 & 0.355 \\
    F--C & 0.254 & 0.217 & 0.292 \\
    E--C & 0.019 & 0.006 & 0.046 \\
    R--T & 0.197 & 0.100 & 0.262 \\
    \hline
  \end{tabular}
\end{table}

\begin{table}[t]
  \centering
  \scriptsize
  \setlength{\tabcolsep}{4pt}
  \caption{Pareto-optimal strategies (not strictly dominated by any other strategy).}
  \label{tab:app_rq2_pareto}
  \begin{tabular}{lll}
    \hline
    Model & Pareto (E/F/C) & Pareto (All metrics) \\
    \hline
    GPT-5-Nano & More Quotes; Statistics & Authoritative; More Quotes; Statistics \\
    GPT-OSS-120B & More Quotes & Fluent; More Quotes; Technical Terms \\
    \hline
  \end{tabular}
\end{table}

\subsection{Heterogeneity by Query Tags}

To study heterogeneity, we slice the test set by seven query tag fields (e.g., User Intent, Answer Type).
For each slice with at least 200 instances, we report the strategy that maximizes $\Delta$C within that slice (Tables~\ref{tab:app_rq3_best_slices_nano}--\ref{tab:app_rq3_best_slices_oss}).
Small slices are noisier and should be interpreted with caution.

\begin{table*}[t]
  \centering
  \scriptsize
  \setlength{\tabcolsep}{3pt}
  \caption{GPT-5-Nano: For each tag slice with $n \ge 200$, the best strategy under $\Delta$C.}
  \label{tab:app_rq3_best_slices_nano}
  \begin{tabular}{llrlrrr}
    \hline
    Tag Field & Tag Value & $n$ & Best Strategy & $\Delta$C & $\Delta$F & $\Delta$E \\
    \hline
    Answer Type & Explanation & 3758 & Statistics & +0.144 & +0.354 & -0.079 \\
    Answer Type & Fact & 568 & Statistics & +0.121 & +0.009 & -0.166 \\
    Answer Type & List & 770 & Statistics & +0.086 & +0.204 & +0.047 \\
    Difficulty Level & Complex & 447 & Statistics & +0.145 & +0.416 & -0.154 \\
    Difficulty Level & Intermediate & 4280 & Statistics & +0.137 & +0.333 & -0.061 \\
    Difficulty Level & Simple & 618 & Statistics & +0.139 & -0.010 & -0.099 \\
    Genre & Arts and Entertainment & 811 & More Quotes & +0.108 & +0.376 & +0.018 \\
    Genre & Books and Literature & 201 & Credible Sources & +0.030 & -0.269 & +0.080 \\
    Genre & Health & 650 & Statistics & +0.163 & +0.845 & +0.005 \\
    Genre & Law and Government & 482 & Statistics & +0.056 & -0.274 & -0.089 \\
    Genre & People and Society & 587 & More Quotes & +0.054 & +0.608 & +0.058 \\
    Genre & Science & 840 & Statistics & +0.268 & +0.424 & -0.135 \\
    Genre & Sports & 308 & More Quotes & +0.166 & +0.562 & +0.101 \\
    Nature of Query & Comparison & 203 & Statistics & +0.207 & +0.852 & -0.103 \\
    Nature of Query & Informational & 4851 & Statistics & +0.133 & +0.278 & -0.069 \\
    Nature of Query & Instructional & 240 & Statistics & +0.154 & +0.350 & -0.146 \\
    Sensitivity & Non-sensitive & 4608 & Statistics & +0.155 & +0.307 & -0.073 \\
    Sensitivity & Sensitive & 745 & Statistics & +0.029 & +0.267 & -0.074 \\
    Specific Topics & Biology & 655 & Statistics & +0.237 & +0.644 & -0.124 \\
    Specific Topics & Not Applicable & 4167 & Statistics & +0.116 & +0.221 & -0.054 \\
    Specific Topics & Physics & 243 & Statistics & +0.259 & +0.346 & -0.107 \\
    User Intent & Learning & 2297 & Statistics & +0.194 & +0.444 & -0.060 \\
    User Intent & Research & 2997 & Statistics & +0.094 & +0.189 & -0.078 \\
    \hline
  \end{tabular}
\end{table*}

\begin{table*}[t]
  \centering
  \scriptsize
  \setlength{\tabcolsep}{3pt}
  \caption{GPT-OSS-120B: For each tag slice with $n \ge 200$, the best strategy under $\Delta$C.}
  \label{tab:app_rq3_best_slices_oss}
  \begin{tabular}{llrlrrr}
    \hline
    Tag Field & Tag Value & $n$ & Best Strategy & $\Delta$C & $\Delta$F & $\Delta$E \\
    \hline
    Answer Type & Explanation & 3758 & More Quotes & +0.147 & +0.601 & +0.154 \\
    Answer Type & Fact & 568 & Statistics & +0.115 & +0.259 & -0.111 \\
    Answer Type & List & 770 & More Quotes & +0.233 & +0.523 & +0.156 \\
    Difficulty Level & Complex & 447 & Statistics & +0.163 & +0.347 & +0.007 \\
    Difficulty Level & Intermediate & 4280 & More Quotes & +0.159 & +0.612 & +0.142 \\
    Difficulty Level & Simple & 618 & More Quotes & +0.123 & +0.455 & +0.005 \\
    Genre & Arts and Entertainment & 811 & More Quotes & +0.113 & +0.443 & +0.136 \\
    Genre & Books and Literature & 201 & More Quotes & -0.025 & +0.159 & +0.139 \\
    Genre & Health & 650 & More Quotes & +0.212 & +1.026 & +0.148 \\
    Genre & Law and Government & 482 & Statistics & +0.197 & +0.100 & -0.075 \\
    Genre & People and Society & 587 & More Quotes & +0.136 & +0.673 & +0.094 \\
    Genre & Science & 840 & Statistics & +0.148 & +0.501 & -0.123 \\
    Genre & Sports & 308 & Statistics & +0.136 & +0.477 & -0.110 \\
    Nature of Query & Comparison & 203 & More Quotes & +0.251 & +0.823 & +0.295 \\
    Nature of Query & Informational & 4851 & More Quotes & +0.144 & +0.580 & +0.123 \\
    Nature of Query & Instructional & 240 & SEO & +0.225 & +0.583 & +0.062 \\
    Sensitivity & Non-sensitive & 4608 & More Quotes & +0.144 & +0.539 & +0.154 \\
    Sensitivity & Sensitive & 745 & More Quotes & +0.174 & +0.807 & -0.005 \\
    Specific Topics & Biology & 655 & Statistics & +0.197 & +0.662 & +0.069 \\
    Specific Topics & Not Applicable & 4167 & More Quotes & +0.144 & +0.558 & +0.117 \\
    Specific Topics & Physics & 243 & More Quotes & +0.185 & +0.354 & +0.185 \\
    User Intent & Learning & 2297 & More Quotes & +0.154 & +0.639 & +0.169 \\
    User Intent & Research & 2997 & More Quotes & +0.146 & +0.528 & +0.106 \\
    \hline
  \end{tabular}

\end{table*}

\subsection{Effect of Retrieval Position}

We analyze how the target document's retrieval position affects performance.
Table~\ref{tab:app_position_baseline_nano} and Table~\ref{tab:app_position_baseline_oss} show that both E and F decrease substantially as the target document appears lower in the retrieved list.
As an example of mitigation at low rank (position=4): for GPT-5-Nano, \emph{Statistics} increases F from 3.755 to 4.308 (+0.554); for GPT-OSS-120B, \emph{More Quotes} increases F from 3.768 to 4.586 (+0.818).

\begin{table}[t]
  \centering
  \scriptsize
  \setlength{\tabcolsep}{4pt}
  \caption{GPT-5-Nano (None baseline): Performance by retrieval position. Pos.=0 means top-ranked.}
  \label{tab:app_position_baseline_nano}
  \begin{tabular}{rrrrr}
    \hline
    Pos. & $n$ & E & F & C \\
    \hline
    0 & 1882 & 6.209 & 7.353 & 5.774 \\
    1 & 839 & 5.744 & 6.397 & 5.543 \\
    2 & 608 & 5.632 & 5.515 & 5.345 \\
    3 & 521 & 5.511 & 4.660 & 5.276 \\
    4 & 1503 & 5.171 & 3.755 & 5.138 \\
    \hline
  \end{tabular}
\end{table}

\begin{table}[t]
  \centering
  \scriptsize
  \setlength{\tabcolsep}{4pt}
  \caption{GPT-OSS-120B (None baseline): Performance by retrieval position. Pos.=0 means top-ranked.}
  \label{tab:app_position_baseline_oss}
  \begin{tabular}{rrrrr}
    \hline
    Pos. & $n$ & E & F & C \\
    \hline
    0 & 1882 & 6.181 & 7.356 & 5.759 \\
    1 & 839 & 5.725 & 6.337 & 5.508 \\
    2 & 608 & 5.558 & 5.556 & 5.413 \\
    3 & 521 & 5.570 & 4.708 & 5.378 \\
    4 & 1503 & 5.138 & 3.768 & 5.105 \\
    \hline
  \end{tabular}
\end{table}

\subsection{Document Feature Changes and Feature--Metric Correlations}

We quantify how each optimization changes document-level features and relate these changes to metric improvements.
Table~\ref{tab:app_feature_deltas_nano} and Table~\ref{tab:app_feature_deltas_oss} summarize average feature deltas relative to the original document.
Table~\ref{tab:app_top_correlations_nano} and Table~\ref{tab:app_top_correlations_oss} list the strongest (absolute) feature--metric correlations within each strategy.
These correlations are intended as interpretability signals and should not be interpreted causally.

\begin{table}[t]
  \centering
  \scriptsize
  \setlength{\tabcolsep}{2pt}
  \caption{GPT-5-Nano: Mean document feature changes (optimized minus original). $\Delta$ch=characters, $\Delta$w=words, $\Delta$uw=unique words, $\Delta$num=numbers, $\Delta$cit=citations, $\Delta$quo=quotes.}
  \label{tab:app_feature_deltas_nano}
  \resizebox{\columnwidth}{!}{
    \begin{tabular}{lrrrrrr}
      \hline
      Strategy & $\Delta$ch & $\Delta$w & $\Delta$uw & $\Delta$num & $\Delta$cit & $\Delta$quo \\
      \hline
      Authoritative & -17.3 & -6.5 & -0.4 & -0.2 & 0.0 & -0.6 \\
      Credible Sources & +473.9 & +67.7 & +37.9 & +1.7 & -0.3 & -0.2 \\
      Fluent & -32.2 & -4.9 & +0.3 & -0.4 & -0.2 & -0.7 \\
      More Quotes & +389.0 & +53.7 & +31.5 & +0.4 & 0.0 & +6.1 \\
      SEO & +245.5 & +31.2 & +21.9 & +0.0 & -0.1 & -0.1 \\
      Simple Language & -58.9 & -7.6 & -2.8 & -0.2 & -0.1 & -0.6 \\
      Statistics & +398.0 & +56.9 & +33.4 & +7.2 & -0.1 & -0.1 \\
      Technical Terms & +31.1 & -2.9 & +3.3 & -0.2 & -0.1 & -0.7 \\
      Unique Words & +9.2 & -3.7 & +3.5 & -0.5 & -0.2 & -0.9 \\
      \hline
    \end{tabular}
  }
\end{table}

\begin{table}[t]
  \centering
  \scriptsize
  \setlength{\tabcolsep}{2pt}
  \caption{GPT-OSS-120B: Mean document feature changes (optimized minus original). $\Delta$ch=characters, $\Delta$w=words, $\Delta$uw=unique words, $\Delta$num=numbers, $\Delta$cit=citations, $\Delta$quo=quotes.}
  \label{tab:app_feature_deltas_oss}
  \resizebox{\columnwidth}{!}{
    \begin{tabular}{lrrrrrr}
      \hline
      Strategy & $\Delta$ch & $\Delta$w & $\Delta$uw & $\Delta$num & $\Delta$cit & $\Delta$quo \\
      \hline
      Authoritative & +112.8 & +12.9 & +11.0 & -0.1 & -0.1 & +0.0 \\
      Credible Sources & +462.6 & +66.1 & +40.7 & +2.5 & -0.3 & +0.4 \\
      Fluent & -41.2 & -7.9 & +3.7 & -0.2 & -0.2 & +0.2 \\
      More Quotes & +487.6 & +66.7 & +40.1 & +2.5 & +0.0 & +9.7 \\
      SEO & +303.5 & +39.1 & +27.4 & +0.0 & -0.2 & +0.0 \\
      Simple Language & -85.8 & -11.4 & -3.1 & -0.1 & -0.2 & +0.0 \\
      Statistics & +326.5 & +45.9 & +28.0 & +7.6 & 0.0 & +0.1 \\
      Technical Terms & +100.4 & +1.0 & +10.9 & -0.1 & -0.1 & +0.0 \\
      Unique Words & +65.3 & -0.7 & +8.7 & -0.4 & -0.2 & +0.1 \\
      \hline
    \end{tabular}
  }
\end{table}

\begin{table}[t]
  \centering
  \scriptsize
  \setlength{\tabcolsep}{4pt}
  \caption{GPT-5-Nano: Top Pearson correlations between feature changes and metric deltas (within each optimization).}
  \label{tab:app_top_correlations_nano}
  \begin{tabular}{llll}
    \hline
    Strategy & Feature & Metric & $r$ \\
    \hline
    Credible Sources & words & F & +0.341 \\
    Credible Sources & unique words & F & +0.332 \\
    Credible Sources & chars & F & +0.327 \\
    More Quotes & unique words & R & +0.300 \\
    More Quotes & words & F & +0.293 \\
    More Quotes & chars & F & +0.284 \\
    More Quotes & unique words & F & +0.278 \\
    Credible Sources & unique words & R & +0.276 \\
    Credible Sources & words & R & +0.267 \\
    More Quotes & words & R & +0.265 \\
    \hline
  \end{tabular}
\end{table}

\begin{table}[t]
  \centering
  \scriptsize
  \setlength{\tabcolsep}{4pt}
  \caption{GPT-OSS-120B: Top Pearson correlations between feature changes and metric deltas (within each optimization).}
  \label{tab:app_top_correlations_oss}
  \begin{tabular}{llll}
    \hline
    Strategy & Feature & Metric & $r$ \\
    \hline
    Authoritative & unique words & T & -0.381 \\
    Fluent & unique words & R & +0.332 \\
    Authoritative & words & T & -0.329 \\
    Authoritative & chars & T & -0.306 \\
    More Quotes & unique words & R & +0.292 \\
    More Quotes & words & R & +0.269 \\
    More Quotes & chars & R & +0.265 \\
    SEO & unique words & R & +0.210 \\
    Credible Sources & unique words & R & +0.192 \\
    More Quotes & quotes & R & +0.185 \\
    \hline
  \end{tabular}
\end{table}

\subsection{Cross-Model Consistency}

We compare strategy rankings between GPT-5-Nano and GPT-OSS-120B using Spearman correlation (Table~\ref{tab:app_rq6_spearman}).
Rankings are highly consistent for E/F/C but substantially less consistent for R/T.

\begin{table}[t]
  \centering
  \scriptsize
  \setlength{\tabcolsep}{6pt}
  \begin{tabular}{lr}
    \hline
    Metric & Spearman \\
    \hline
    Exposure (E) & 0.917 \\
    Faithful Credit (F) & 0.883 \\
    Causal Impact (C) & 0.917 \\
    Readability \& Structure (R) & 0.533 \\
    Trustworthiness \& Safety (T) & 0.500 \\
    \hline
  \end{tabular}
  \caption{Cross-model Spearman rank correlation of strategy ranking (deltas vs. None).}
  \label{tab:app_rq6_spearman}
\end{table}

\end{document}